\documentclass[sigconf,nonacm]{acmart}

\usepackage{bbm}
\usepackage{enumitem}

\begin{document}

\title{TERRA: Task-Oriented Edge AI for Regional Fire Detection with Identity-Free Reporting}



\author{Mohammad Kazemi}
\affiliation{%
\department{Department of Computing}
 \institution{Imperial College London}
   \city{London}
  \country{United Kingdom}
 }
\email{mohammad.kazemi@imperial.ac.uk}

\author{Julie A. McCann}
\affiliation{%
\department{Department of Computing}
 \institution{Imperial College London}
   \city{London}
  \country{United Kingdom}
 }
\email{j.mccann@imperial.ac.uk}









\begin{abstract}
Distributed fire-monitoring systems may involve many sensing devices reporting over a shared uplink, even though the operational objective is often to identify regions exhibiting evidence of fire rather than to recover device-specific measurements. We introduce \emph{TERRA} (\emph{Task-Oriented Edge Representation and Regional Aggregation}), an edge-AI framework that combines task-oriented local inference, identity-free region-tagged reporting, and permutation-invariant regional evidence aggregation. TERRA supports two reporting modes: \emph{TERRA-Score}, in which each device communicates a quantized local fire probability, and \emph{TERRA-Feature}, in which it communicates a compact learned representation for receiver-side aggregation using a permutation-invariant DeepSets model. Using all 15 numerical sensing modalities of the distributed indoor EN54 fire-alarm dataset as a proof-of-concept testbed, we evaluate regional inference for an episode-level fire-scenario target under repetition-disjoint three-fold cross-validation. We study representation size, task-payload quantization, report availability, and robustness to missed and false recovered messages. With nine reports and perfect message recovery, TERRA-Feature achieves mean AUPRC \(0.990\) using a 24-bit task payload, while TERRA-Score achieves \(0.987\) using only 6 bits; both attain an MDR of \(1.81\%\) at FAR \(\leq1\%\). TERRA-Feature achieves slightly higher AUPRC under reliable recovery, whereas the more compact TERRA-Score is less sensitive to increasing false-message contamination.
\end{abstract}


\maketitle

\section{Introduction}
\label{sec:introduction}

Wildfires threaten ecosystems, infrastructure, and communities, and their impact can be reduced when emerging incidents are detected before they develop into large-scale events. Internet of Things (IoT) systems can complement satellite imagery, aerial surveillance, watchtowers, and public reports by providing persistent local monitoring through heterogeneous sensing modalities such as temperature, humidity, particulate matter, gas concentrations, and optical measurements \cite{Chan2024WildfireIoTSurvey,Sasmita2018LoRaFire,Freira2024WildfireIoT,Herring2022UndergroundLoRa}. In remote or difficult-to-access environments, however, such systems must operate under limited energy, bandwidth, and infrastructure availability while remaining scalable when many sensors report concurrently.

Conventional monitoring architectures commonly recover measurements, features, or alarm decisions as individually addressed device reports. This is appropriate when device identities or specific observations are needed, but can introduce unnecessary addressing and access coordination when the objective is to identify regions showing evidence of an event. Sensors within the same region may observe different manifestations of the same regional state, making their collective task-relevant evidence more important than their persistent identities.

We introduce \emph{TERRA} (\emph{Task-Oriented Edge Representation and Regional Aggregation}), an edge-AI framework that combines task-oriented local inference, identity-free region-tagged reporting, and permutation-invariant regional evidence aggregation.
Each active device processes its multimodal temporal observations locally and communicates a compact task payload, along with a geographic region tag and a nonce, without requiring a persistent transmitter identity. TERRA supports two reporting modes: \emph{TERRA-Score}, in which the device communicates a quantized local fire probability that is averaged across recovered reports, and \emph{TERRA-Feature}, in which it communicates a compact learned feature vector that is fused at the receiver using a permutation-invariant DeepSets model \cite{Zaheer2017DeepSets}. 
For identity-free access, the present system model adopts unsourced random access (URA), in which active devices share a common codebook and the receiver recovers an unordered list of transmitted messages rather than their source identities.

The two modes expose a task-specific communication tradeoff. TERRA-Score communicates the completed local binary inference using very few bits, whereas TERRA-Feature retains richer intermediate information for receiver-side processing at a larger payload. This creates a tradeoff between communication efficiency and the flexibility of receiver-side inference, which we evaluate under both reliable and imperfect message recovery.

\subsection{Related Work and Positioning}
\label{sec:related_work}

IoT-based fire and wildfire monitoring systems employ distributed environmental sensors and wireless networks to support early detection and alarm delivery \cite{Chan2024WildfireIoTSurvey,Sasmita2018LoRaFire,Freira2024WildfireIoT,Herring2022UndergroundLoRa}. Edge-intelligence methods reduce communication by processing observations near their sources and transmitting local decisions or intermediate representations \cite{Zhou2019EdgeIntelligence,Teerapittayanon2017DDNN}, while task-oriented communication further tailors the communicated information to a downstream inference objective rather than source reconstruction \cite{Gunduz2023BeyondBits,Shao2022TaskOriented}. TERRA combines these ideas with region-wise evidence fusion: the required output is a regional event decision, and devices communicate either task-specific scores or learned representations rather than complete sensor records.

Over-the-air computation can directly exploit waveform superposition to recover functions such as sums or averages of distributed observations \cite{Liu2020AirComp,Sahin2023AirCompSurvey}, and could provide an alternative way to implement simple score aggregation. TERRA instead recovers discrete task reports before regional aggregation, allowing the same framework to accommodate explicit region tags and both scalar scores and learned feature vectors. We model identity-free reporting through URA, in which devices share a common codebook and the receiver recovers transmitted messages without persistent source identities; \cite{Ozates2026URASurvey} provides a comprehensive overview of this access paradigm. In TERRA, the recovered messages are an intermediate representation of distributed evidence rather than the final application output: they are grouped by region and subsequently fused to obtain the regional decision.

\subsection{Contributions}
\label{sec:contributions}

The main contributions of this work are:
\begin{itemize}[leftmargin=*]
    \item We introduce TERRA, a task-oriented distributed edge-AI architecture in which sensing devices transform local observations into compact task reports, transmit them without persistent device identities using explicit region tags, and enable permutation-invariant fusion of the recovered reports into regional decisions. The framework supports complementary TERRA-Score and TERRA-Feature operating modes that communicate completed local decisions or intermediate learned representations, respectively.
    
    \item We develop a staged evaluation that separates representation design, report availability, task-payload quantization, and message-recovery errors, using a compact nonlearned Handcrafted-Stats pipeline as a benchmark.

    \item Using the distributed indoor EN54 fire dataset as a proof-of-concept testbed, we select a four-dimensional learned representation for TERRA-Feature and use the same representation dimension in the sensor-side models of both TERRA modes. With nine reports and perfect message recovery, TERRA-Feature achieves mean AUPRC \(0.990\) using a 24-bit task payload, while TERRA-Score achieves \(0.987\) using only 6 bits; both attain an MDR of \(1.81\%\) at FAR \(\leq1\%\), compared with an AUPRC of \(0.971\) and MDR of \(11.42\%\) for the 30-bit Handcrafted-Stats benchmark.
    
    \item We characterize how the choice between communicating intermediate learned representations and completed local decisions affects regional inference over a broad range of missed and false recovered-message rates. TERRA-Feature performs best under reliable recovery, whereas the substantially more compact TERRA-Score is less sensitive to increasing false-message contamination, exposing a communication--inference tradeoff that is hidden when edge representation and regional aggregation are considered separately.
\end{itemize}


\section{TERRA System Model and Framework}
\label{sec:terra}

\begin{figure*}[t]
    \centering
    \includegraphics[trim={0cm 0.2cm 0cm 0cm}, clip, width=0.85\linewidth]{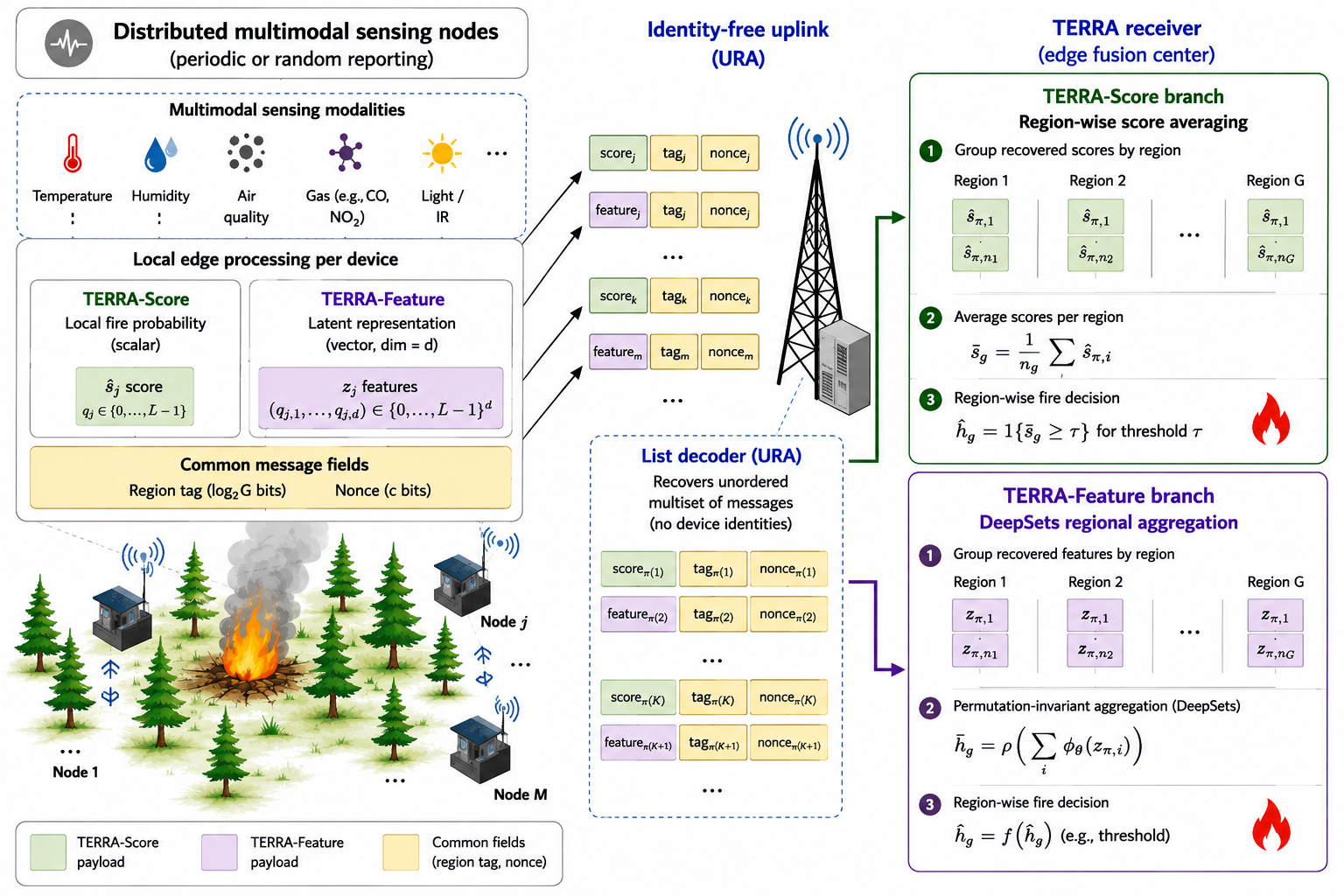}
    \caption{TERRA architecture for region-wise inference over an identity-free random access channel. Devices transmit either a quantized local score (TERRA-Score) or a learned representation (TERRA-Feature), together with a region tag and a nonce; recovered reports are grouped by region and aggregated using score averaging or DeepSets, respectively.}
    \Description{TERRA architecture with distributed multimodal sensing nodes performing local score or feature extraction, transmitting region-tagged reports over an identity-free URA uplink, and a receiver grouping recovered reports by region for score averaging or permutation-invariant DeepSets aggregation.}
    \label{fig:terra_architecture}
\end{figure*}

We consider a distributed monitoring system in which spatially deployed sensing devices observe a common environment and communicate with a central receiver through a shared uplink. The application objective is a region-wise event map rather than a collection of device-specific measurements: the receiver should determine which geographic regions exhibit evidence of fire without requiring persistent transmitter identities. TERRA addresses this objective by combining sensor-side edge inference, compact identity-free reporting, and region-wise evidence fusion. Figure~\ref{fig:terra_architecture} illustrates the two reporting modes considered in this work.

\subsection{Spatial Sensing and Edge Reporting Modes}
\label{sec:terra_edge}

Let the monitored area be partitioned into \(G\) nonoverlapping geographic regions indexed by \([G]\triangleq\{1,\ldots,G\}\), and let \(\mathcal{K}\triangleq\{1,\ldots,K\}\) denote the deployed sensing devices. Device \(k\in\mathcal{K}\) is associated with a region index \(g_k\in[G]\), assumed to be known locally through deployment configuration, positioning information, or another localization mechanism, while \(y_g\in\{0,1\}\) denotes the state of region \(g\), with \(y_g=1\) indicating the presence of the target event.


At each reporting opportunity, device \(k\) forms a temporal multimodal observation \(\mathbf{X}_k\in\mathbb{R}^{C\times W}\) from its most recent \(W\) sensor samples, where \(C\) is the number of sensing modalities. The regional target \(y_g\) denotes the application-level state associated with that reporting opportunity. Reporting opportunities occur periodically or randomly according to the access policy, and a device may or may not transmit at a given opportunity.

Let \(a_k\in\{0,1\}\) indicate whether device \(k\) transmits at the current reporting opportunity and define the active set \(\mathcal{A}\triangleq\{k\in\mathcal{K}:a_k=1\}\). The number of transmitting devices associated with region \(g\) is \(K_g=\sum_{k\in\mathcal{K}}a_k\mathbbm{1}\{g_k=g\}\). Activity in the baseline system is independent of the local observation and event state, so the absence of a report is not itself interpreted as evidence of a non-fire state.

TERRA provides two alternative reporting configurations. In \emph{TERRA-Score}, the edge model completes the local binary inference task and produces a fire probability
\begin{equation}
s_k=f_{\boldsymbol{\theta}_{\mathrm{s}}}(\mathbf{X}_k)\in[0,1],
\label{eq:terra_score}
\end{equation}
which is communicated as a compact scalar task payload. In \emph{TERRA-Feature}, the edge model instead extracts an intermediate representation
\begin{equation}
\mathbf{z}_k=f_{\boldsymbol{\theta}_{\mathrm{f}}}(\mathbf{X}_k)\in\mathbb{R}^{d},
\label{eq:terra_feature}
\end{equation}
which preserves information for receiver-side regional inference. A device uses one of these reporting modes at a time: TERRA-Score communicates the completed local inference as a compact scalar, whereas TERRA-Feature communicates an intermediate learned representation for receiver-side inference at a larger task payload.

\subsection{Task-Oriented Messages and Identity-Free Access}
\label{sec:terra_message}

Let \(r\in\{\mathrm{S},\mathrm{F}\}\) denote the selected reporting mode and define \(t_k^{(\mathrm{S})}=s_k\) and \(t_k^{(\mathrm{F})}=\mathbf{z}_k\). Before transmission, the task quantity is converted to a finite-bit description
\begin{equation}
\mathbf{q}_k^{(r)}
=
Q^{(r)}\!\left(t_k^{(r)}\right)
\in\{0,1\}^{B_{\mathrm{t}}^{(r)}},
\label{eq:task_quantization}
\end{equation}
where \(B_{\mathrm{t}}^{(r)}\) denotes the task-payload length. Device \(k\) constructs the message
\begin{equation}
\mathbf{m}_k^{(r)}
=
\gamma(g_k)\,\Vert\,\mathbf{q}_k^{(r)}\,\Vert\,\mathbf{u}_k,
\qquad
B^{(r)}
=
B_g+B_{\mathrm{t}}^{(r)}+B_u,
\label{eq:terra_message}
\end{equation}
where \(\gamma(g_k)\in\{0,1\}^{B_g}\) encodes the geographic region, \(\mathbf{u}_k\in\{0,1\}^{B_u}\) is a fresh randomly generated nonce for the report, and \(\Vert\) denotes concatenation. The region field contains only the spatial information required by the task and does not identify the device. The nonce likewise carries no persistent device identity; its role is to distinguish complete messages whose region and quantized task fields would otherwise be identical.

We instantiate the identity-free access abstraction using unsourced random access (URA), in which active devices use a common codebook and the receiver recovers the transmitted messages without resolving persistent transmitter identities \cite{Ozates2026URASurvey}. Here, \emph{identity-free} refers to these access and inference semantics: no persistent device identifier is required in the task message or for regional aggregation.
A generic multiple-access observation can be represented as
\begin{equation}
\mathbf{y}
=
\sum_{k\in\mathcal{A}}
h_k\mathcal{C}^{(r)}\!\left(\mathbf{m}_k^{(r)}\right)
+\mathbf{w},
\qquad
\widehat{\mathcal{M}}
=
\mathcal{D}_{\mathrm{URA}}^{(r)}(\mathbf{y}),
\label{eq:ura_channel}
\end{equation}
where \(\mathcal{C}^{(r)}\) is the common encoder, \(h_k\) denotes the effective channel coefficient, \(\mathbf{w}\) represents channel noise, and \(\mathcal{D}_{\mathrm{URA}}^{(r)}\) denotes a list decoder that returns the unordered recovered-message list \(\widehat{\mathcal{M}}\).

\subsection{Region-Wise Task Aggregation}
\label{sec:terra_aggregation}

Each recovered message \(\widehat{\mathbf{m}}_j\in\widehat{\mathcal{M}}\) is parsed into a decoded region index \(\widehat{g}_j\), quantized task field \(\widehat{\mathbf{q}}_j\), and nonce, after which the nonce is discarded. For region \(g\), define its recovered report indices as
\begin{equation}
\widehat{\mathcal{I}}_g
=
\{j:\widehat{g}_j=g,\ \widehat{\mathbf{m}}_j\in\widehat{\mathcal{M}}\},
\qquad
\widehat{K}_g=|\widehat{\mathcal{I}}_g|.
\label{eq:regional_indices}
\end{equation}

For TERRA-Score, the reconstructed local probability is
\(\widehat{s}_j=(Q^{(\mathrm{S})})^{-1}(\widehat{\mathbf{q}}_j)\), and the regional probability is obtained by averaging the recovered scores,
\begin{equation}
\widehat{p}_g^{(\mathrm{S})}
=
\frac{1}{\widehat{K}_g}
\sum_{j\in\widehat{\mathcal{I}}_g}
\widehat{s}_j,
\qquad \widehat{K}_g>0.
\label{eq:score_aggregation}
\end{equation}
Thus, TERRA-Score performs task inference locally and uses the receiver only to combine reported probabilities within each region.

For TERRA-Feature, the reconstructed report is
\(\widehat{\mathbf{z}}_j=(Q^{(\mathrm{F})})^{-1}(\widehat{\mathbf{q}}_j)\), and the regional evidence is processed using a DeepSets architecture \cite{Zaheer2017DeepSets},
\begin{equation}
\overline{\mathbf{h}}_g
=
\frac{1}{\widehat{K}_g}
\sum_{j\in\widehat{\mathcal{I}}_g}
\phi_{\boldsymbol{\psi}_{\phi}}
\!\left(\widehat{\mathbf{z}}_j\right),
\qquad
\widehat{p}_g^{(\mathrm{F})}
=
\sigma\!\left(
\rho_{\boldsymbol{\psi}_{\rho}}
\!\left(\overline{\mathbf{h}}_g\right)
\right),
\label{eq:feature_aggregation}
\end{equation}
where \(\phi_{\boldsymbol{\psi}_{\phi}}\) is a learned neural network that transforms each individual report, \(\rho_{\boldsymbol{\psi}_{\rho}}\) is a learned neural network that maps the pooled regional representation to a scalar logit, and \(\boldsymbol{\psi}_{\phi}\) and \(\boldsymbol{\psi}_{\rho}\) denote their trainable parameters. The averaging operation makes the receiver invariant to the ordering of recovered reports; in a batched implementation, a mask simply ensures that padded positions do not contribute to this mean. The function \(\sigma(x)=1/(1+e^{-x})\) is the logistic sigmoid, producing the regional fire probability.

If no report is assigned to a region, a predefined default output is required; in the experiments, we set the regional fire probability to zero. A binary alarm can subsequently be obtained by thresholding \(\widehat{p}_g\), while the evaluation procedure and operating-point metrics are described in Section~\ref{sec:experimental_setup}.

\subsection{Message-Recovery Errors}
\label{sec:terra_errors}

Let $\mathcal{M}
=
\{\mathbf{m}_k^{(r)}:k\in\mathcal{A}\}$
denote the transmitted full-message set and \(\widehat{\mathcal{M}}\) the recovered list. We characterize message recovery by
\begin{equation}
P_{\mathrm{MD}}
=
\mathbb{E}\!\left[
\frac{|\mathcal{M}\setminus\widehat{\mathcal{M}}|}
{|\mathcal{M}|}
\right],
\qquad
P_{\mathrm{FA}}
=
\mathbb{E}\!\left[
\frac{|\widehat{\mathcal{M}}\setminus\mathcal{M}|}
{|\widehat{\mathcal{M}}|}
\right],
\label{eq:md_fa}
\end{equation}
where the false-alarm term is defined as zero when the recovered list is empty. A missed detection removes genuine regional evidence, whereas a false alarm introduces a decoded message that was not transmitted and is assigned to a region according to its decoded tag. A decoding error that converts one transmitted message into another therefore appears at the recovered-list level as a missed genuine message together with a false decoded message, so the abstraction does not require separate payload-bit or region-tag error probabilities.

\section{Experimental Methodology}
\label{sec:experimental_setup}

We first evaluate the application pipeline under perfect message recovery to isolate the effects of representation size, report availability, and task-payload quantization. We then introduce controlled missed and false message recoveries to characterize communication robustness. We train all sensor-side models and regional receivers only on development data under the repetition-level cross-validation procedure described below. We then freeze the resulting local scores, learned representations, and regional receivers before the report-availability, quantization, and message-recovery experiments, so these evaluations do not involve quantization-aware fine-tuning or adaptation to communication errors.

\subsection{Dataset, Windowing, and Cross-Validation}
\label{sec:dataset_cv}

We use the distributed indoor EN54 fire-alarm dataset \cite{Vorwerk2023Distance,Vorwerk2025Dataset}. After preprocessing, the data used in our experiments contain 304,920 timestamped node-level measurements collected from nine synchronized multisensor nodes during controlled fire, nuisance, and background experiments. EN54 provides the synchronized multi-node, multimodal observations needed to evaluate the proposed distributed inference architecture and is therefore used as a controlled proof-of-concept fire-detection testbed rather than direct validation of an outdoor wildfire deployment. We retain all 15 numerical sensing modalities available at the nodes: carbon dioxide, carbon monoxide, hydrogen, humidity, particulate-matter measurements corresponding to approximately \(0.5\), \(1.0\), \(2.5\), \(4.0\), and \(10~\mu\mathrm{m}\), typical particle size, total particulate matter, temperature, ultraviolet intensity, volatile organic compounds (VOC), and raw VOC measurements.

The recordings are divided into \emph{episodes} corresponding to contiguous runs of the same experimental scenario, with a new episode also introduced when the gap between consecutive measurements exceeds 30~s. Within each sensor trace and episode, measurements are partitioned into consecutive, non-overlapping windows of \(W=12\) samples, corresponding to approximately two minutes at the nominal 10-s sampling interval. We use \(W=12\) as a fixed temporal-context design point and do not optimize the window duration in this study. Each window endpoint defines one reporting opportunity, so a window is the local temporal observation available to a sensor at that opportunity, whereas an episode is the longer experimental sequence from which multiple such windows may be obtained. Windows never cross sensor or episode boundaries.

The controlled Cable, Candles, Lunts, and Wood recordings constitute the fire scenarios, while Background and nuisance experiments constitute the non-fire scenarios. The dataset provides two fire labels that play different roles in our pipeline. The sensor-side classifier is supervised using the sample-level control fire label (\texttt{fire\_label\_control}), which can change during an experimental run; consequently, each local \(W\)-sample window is assigned the control label of its final sample. For regional evaluation, we use the binary scenario-level fire label (\texttt{fire\_label}), which is synchronized across sensing nodes and constant within each derived episode. The regional AUPRC and MDR values therefore quantify recognition of whether a synchronized regional example belongs to a controlled fire scenario, rather than ignition-time estimation or alarm-delay performance. The sensor-side auxiliary task provides temporally localized supervision, whereas the final regional task evaluates episode-level fire-scenario recognition.


We normalize each sensing modality using statistics estimated exclusively from clean Background measurements in the corresponding training data. We normalize each physical sensor separately to reduce sensor-specific offsets and scale differences; when fewer than 50 Background samples are available for an individual sensor, we use the corresponding global Background statistics estimated from the same training data instead. Validation and test measurements never contribute to these normalization statistics.

A \emph{repetition} denotes a separate experimental execution of the same controlled scenario. Each of the four fire scenarios contains three repetitions, enabling three repetition-disjoint outer cross-validation folds. In each outer fold, we hold out one repetition of every fire scenario for testing, and the remaining repetitions form part of the development data; we assign non-fire episodes across folds so every episode appears in the test set exactly once. Within each outer development set, we use inner train--validation partitions to select the training duration based on validation AUPRC, after which we retrain the final model from a new initialization on the full outer development set for the selected number of epochs. This construction prevents temporal windows originating from the same experimental episode from appearing in both development and test data.

Regional examples are then formed by synchronizing windows from the nine sensing nodes in the same episode and reporting opportunity, yielding nine candidate sensor reports for each regional decision. Thus, each EN54-derived regional example represents multisensor fusion within one controlled sensing environment; the dataset does not provide multiple independently measured geographic regions.
The three outer folds contain 1540/1265, 2473/332, and 1597/1208 development/test regional examples, respectively.

\subsection{Edge Models, Regional Receiver, and Representation Selection}
\label{sec:models}


Both TERRA modes use the same sensor-side neural architecture. Its direct observation is the normalized \(15\times12\) multimodal window described above. Within the encoder, each sensing-modality trajectory is deterministically transformed into values relative to the first sample, first temporal differences, within-window standardized temporal shape, and three temporal statistics: standard deviation, endpoint change, and linear slope. These quantities are computed from the normalized sensor window inside the sensor-side processing pipeline rather than supplied as additional measurements. The resulting feature vector is processed by a multilayer perceptron with two 64-unit hidden layers and GELU activations to produce a \(d\)-dimensional embedding, followed by a binary-classification head trained with cross-entropy. The sigmoid output of this head provides the local probability used by TERRA-Score, whereas TERRA-Feature communicates the learned embedding itself.

We train the sensor-side models with AdamW using a learning rate of \(10^{-3}\), weight decay \(10^{-4}\), batch size 256, and a maximum of 30 epochs, with early stopping based on validation AUPRC (patience 6). All main experiments use random seed 42.
For TERRA-Feature, we evaluate
$d\in\{1,2,4,8,16,24,35,48,64\}$
and select the smallest dimension whose mean regional AUPRC is within \(0.005\) of the best observed value and whose mean MDR at FAR \(\leq1\%\) is within \(0.02\) in absolute value of the best observed value. This rule selects \(d=4\). Because this dimension study reuses the same three outer folds, we treat it as a development ablation rather than an additional independent estimate of generalization.

After selecting \(d=4\), the corresponding sensor-side model is used for \emph{both} TERRA modes in all main communication experiments. TERRA-Score therefore communicates the scalar classifier probability produced from the same four-dimensional internal representation that TERRA-Feature communicates directly. This matching avoids confounding the comparison between the two reporting modes with differences in sensor-side representation size or model complexity: the principal difference is whether the completed local inference or the intermediate learned representation is transmitted.

The TERRA-Feature regional receiver uses a DeepSets architecture \cite{Zaheer2017DeepSets}, with a report-level network \(\phi\) having hidden/output dimensions \(32/16\) and a receiver network \(\rho\) with hidden dimension \(16\); GELU activations and dropout \(0.15\) are used throughout.
A learned report-level network independently transforms each recovered feature vector, masked mean pooling combines the available transformed reports, and a second learned network maps the resulting order-invariant regional representation to a fire probability. The mask excludes padded entries when variable numbers of reports are processed in a batch. Regional models again use validation AUPRC to select the training duration before being retrained from a new initialization on the complete outer development set.



As a nonlearned edge benchmark, Handcrafted-Stats forms reports from a predefined nested sequence of per-modality temporal statistics: mean, standard deviation, slope, final value, maximum, minimum, and endpoint change. We apply the same near-saturation selection principle used for the learned representation. Its regional predictions are produced by a separately trained DeepSets receiver with the same architecture and training protocol as TERRA-Feature.

\subsection{Report Availability and Task-Payload Quantization}
\label{sec:quantization}

We first assume perfect message recovery and vary only the amount of regional evidence available to the receiver. Starting from each synchronized nine-report regional example, we retain exactly
$K\in\{1,2,3,5,7,9\}$
reports. We evaluate ten independently generated report subsets for each \(K\), using identical subset realizations for all competing schemes. This experiment therefore measures sensitivity to report availability without introducing missed or false decoded messages.

We then apply task-payload quantization post hoc to the frozen models. For TERRA-Score, the local probability \(s\in[0,1]\) is uniformly quantized using
$b\in\{1,2,3,4,6,8\}$
bits with reconstruction levels including the endpoints of the interval. For \(L=2^b\) levels, the reconstruction can be written as
\begin{equation}
\widehat{s}
=
\frac{q}{L-1},
\qquad
q\in\{0,\ldots,L-1\},
\label{eq:score_quantizer}
\end{equation}
where \(q\) is obtained by rounding the scaled probability to the nearest valid index.

Unlike TERRA-Score, for which endpoint reconstruction preserves the meaningful probability limits \(0\) and \(1\), TERRA-Feature and Handcrafted-Stats treat their standardized components as clipped continuous variables and therefore use uniform mid-rise scalar quantization. Each component is standardized using the development-set mean and standard deviation of its corresponding frozen regional receiver and clipped to \([-4,4]\) before quantization. With \(L=2^b\) reconstruction levels and
\(\Delta=8/L\),
\begin{equation}
\widehat{x}
=
-4+\left(q+\frac{1}{2}\right)\Delta,
\qquad
q\in\{0,\ldots,L-1\},
\label{eq:midrise_quantizer}
\end{equation}
where again \(b\in\{1,2,3,4,6,8\}\). We use no per-component range optimization, unequal bit allocation, quantization-aware training, or receiver fine-tuning.

We use the resulting rate--performance curves to select compact near-saturated operating points for the subsequent message-recovery study. These are a 6-bit scalar task payload for TERRA-Score, \(6\) bits per component for the four-dimensional TERRA-Feature representation, giving \(24\) task bits in total, and \(1\) bit per component for the 30-dimensional Handcrafted-Stats representation, giving \(30\) task bits. These values refer only to the task field \(B_{\mathrm t}\); the region tag and nonce are accounted for separately in the system model.

\subsection{Message-Recovery Sensitivity Study and Metrics}
\label{sec:ura_study}

The nonideal communication study characterizes downstream regional inference directly as a function of the list-level missed-detection and false-alarm probabilities \((P_{\mathrm{MD}},P_{\mathrm{FA}})\), without assuming a particular physical-layer code, SNR, or blocklength.
For each target-region example, the nine synchronized EN54 nodes provide up to nine genuine reports. We embed this target region in a synthetic address space of \(G=16\) valid region tags and evaluate \(P_{\mathrm{FA}}\in\{0,0.01,0.05,0.1,0.2,0.3,0.4,0.5,0.6\}\), \(P_{\mathrm{MD}}\in\{0,0.1,\ldots,0.9\}\), with 10 Monte Carlo realizations at each of the 90 operating points. The \(G=16\) regions define a synthetic regional address space for the message-recovery study rather than 16 physically distinct EN54 test areas: we evaluate task performance on the EN54-derived target-region examples, while the remaining regions are represented through their aggregate recovered-message counts when constructing the global recovered list.

Each transmitted full message is independently removed with probability \(P_{\mathrm{MD}}\). Let \(T\) denote the number of genuine messages remaining globally after these missed detections. For \(T>0\), the desired real-valued false-message count is \(F^\star=P_{\mathrm{FA}}T/(1-P_{\mathrm{FA}})\). Since the decoded-list size must be an integer, we randomly choose between \(\lfloor F^\star\rfloor\) and \(\lceil F^\star\rceil\), with the mixing probability chosen so that \(\mathbb{E}[F/(T+F)\mid T]=P_{\mathrm{FA}}\).

Each false decoded message is assigned a uniformly random valid region tag, so target-region contamination corresponds to binomial thinning of the global false-message count with probability \(1/G\). Its task field is sampled independently from the valid quantized task words of the corresponding reporting scheme, and a fresh synthetic nonce ensures that the complete false message was not transmitted. An incorrectly recovered full message is therefore represented at the list level by a missed genuine message and a false decoded message, without separate payload- or region-tag-error probabilities. The uniform tag and task-word distributions are modeling choices for this application-layer sensitivity study.

AUPRC is the primary metric because of class imbalance, with AUROC reported secondarily. We also report the minimum MDR attainable at FAR \(\leq1\%\), and secondarily at FAR \(\leq5\%\); test-set thresholds are not used for model selection. For stochastic experiments, metrics are first averaged across realizations within each outer fold and then equally across the three folds. Given the three outer folds and their reuse for operating-point ablations, we interpret the results as repetition-disjoint engineering estimates rather than formal significance tests.


\section{Results and Discussion}
\label{sec:results}

\subsection{Compact Edge Representations}
\label{sec:results_representation}

We first examine how compactly we can express the learned sensor representation before introducing task-payload quantization or message-recovery errors\footnote{The implementation and experiment configurations are available at
\url{https://github.com/kazemi-mohammad/TERRA}.}. Increasing the TERRA-Feature embedding dimension from \(d=1\) to \(d=4\) clearly improves regional discrimination, after which the gains largely saturate. The mean regional AUPRC increases from \(0.983\) at \(d=1\) and \(0.986\) at \(d=2\) to \(0.990\) at \(d=4\), while the best observed value over \(d\in\{1,2,4,8,16,24,35,48,64\}\) is \(0.993\) at \(d=35\). The corresponding MDR at FAR \(\leq1\%\) is \(1.81\%\) at both \(d=4\) and \(d=35\). Under the predefined near-saturation rule, we therefore select \(d=4\), reducing the transmitted feature dimension by approximately \(89\%\) relative to \(d=35\) while retaining nearly all regional detection performance.

After this selection, we use the same \(d=4\) sensor-side model for both TERRA modes in all subsequent comparisons. The two modes therefore match in sensor-side architecture, representation size, and training procedure, and differ at the reporting interface: TERRA-Score transmits the scalar probability produced by the classifier attached to the \(d=4\) representation, whereas TERRA-Feature transmits the four-dimensional representation itself for receiver-side DeepSets inference. This makes the subsequent comparison a direct tradeoff between communicating a completed local decision and communicating the intermediate information that forms that decision.

We apply the same compaction principle to the nonlearned benchmark. Starting from seven temporal statistics per sensing modality, the predefined nested ablation identifies the per-window mean and standard deviation as the strongest compact configuration, yielding a 30-dimensional Handcrafted-Stats report with a mean unquantized regional AUPRC of \(0.946\). Adding slope, final value, extrema, and endpoint change does not consistently improve the evaluated nested sequence. The final comparison therefore contains three pipelines: TERRA-Score with scalar score averaging, TERRA-Feature with a four-dimensional learned report and a DeepSets receiver, and Handcrafted-Stats with a 30-dimensional mean--standard-deviation report and a separately trained DeepSets receiver using the same architecture and training protocol as the TERRA-Feature receiver.

\subsection{Payload Efficiency and Report Availability}
\label{sec:results_payload}

\begin{figure*}[t]
    \centering
    \includegraphics[trim={0cm 0.35cm 0cm 0cm}, clip, width=0.8\textwidth]{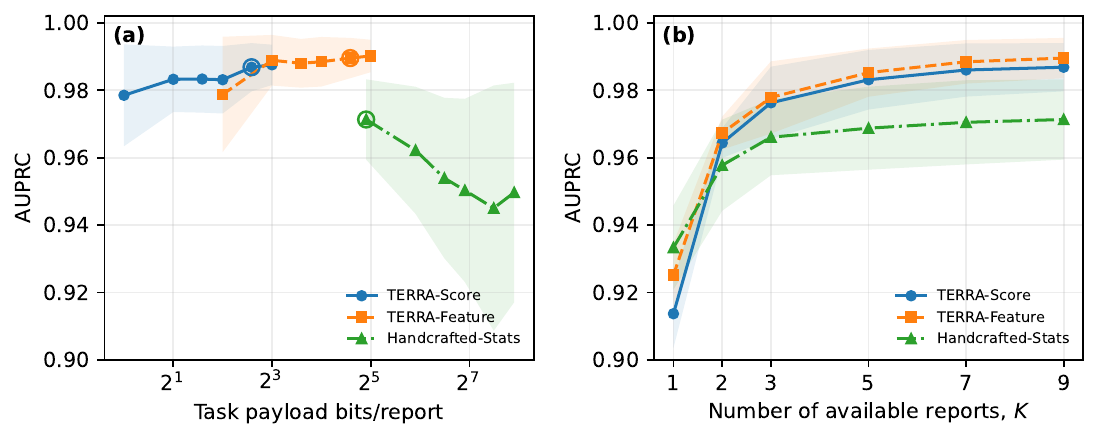}
    \caption{Communication--performance tradeoffs under perfect message recovery. (a) Mean regional AUPRC versus task-payload length with all nine reports available; open markers indicate the selected operating points. (b) Mean regional AUPRC versus the number of available reports using 6-bit TERRA-Score, 24-bit TERRA-Feature, and 30-bit Handcrafted-Stats reports. Shaded regions indicate \(\pm1\) standard deviation across the three outer folds after averaging report-subset realizations within each fold.}
    \Description{Two plots showing regional AUPRC under perfect message recovery. TERRA-Score and TERRA-Feature approach an AUPRC of 0.99 with compact selected payloads, and both improve as the number of available reports increases.}
    \label{fig:payload_reports}
\end{figure*}

Figure~\ref{fig:payload_reports}(a) compares regional AUPRC against the transmitted task-payload length under perfect message recovery with all nine reports available. TERRA-Score reaches a near-saturated operating point using only a few bits: its mean AUPRC is \(0.9868\) with 6 bits and \(0.9876\) with 8 bits, so we retain the 6-bit configuration. TERRA-Feature similarly reaches \(0.9896\) using 6 bits per component, corresponding to a 24-bit task payload for its four-dimensional report, compared with \(0.9902\) at 8 bits per component. Thus, both modes tolerate substantial post-training quantization without quantization-aware training, while TERRA-Score requires substantially fewer task bits because it communicates only the completed local binary inference.

The Handcrafted-Stats benchmark behaves differently. Its most compact evaluated configuration uses one bit for each of the 30 standardized mean and standard-deviation components, yielding a 30-bit task payload and mean AUPRC \(0.9713\), compared with approximately \(0.946\) for the corresponding unquantized representation. With symmetric one-bit mid-rise quantization, each component effectively indicates whether the standardized statistic lies above or below its development-set reference mean, and diagnostic analysis confirms that the resulting binary reports remain diverse and class-informative rather than collapsing to constant patterns. We therefore retain this 30-bit configuration, interpreting the improvement as an empirical regularization effect rather than as evidence that one-bit quantization is intrinsically preferable.

\begin{table}[t]
\centering
\caption{Selected reporting operating points under perfect message recovery with \(K=9\). Report dimension refers to the transmitted task representation; task bits exclude the region tag and nonce.}
\label{tab:operating_points}
\setlength{\tabcolsep}{1pt}
\begin{tabular}{lcccc}
\toprule
Method & Report dim. & Task bits & AUPRC & MDR@FAR$\leq1\%$ \\
\midrule
TERRA-Score       & 1  & 6  & 0.9868 & 1.81\% \\
TERRA-Feature     & 4  & 24 & 0.9896 & 1.81\% \\
Handcrafted-Stats & 30 & 30 & 0.9713 & 11.42\% \\
\bottomrule
\end{tabular}
\end{table}


Table~\ref{tab:operating_points} highlights the resulting communication--inference tradeoff. With matched \(d=4\) sensor-side models, TERRA-Feature achieves a slightly higher clean AUPRC than TERRA-Score, \(0.9896\) versus \(0.9868\), while both attain the same MDR of \(1.81\%\) at FAR \(\leq1\%\). TERRA-Score obtains this performance with only 6 task bits rather than 24, making their distinction more explicit: Feature preserves additional intermediate information for regional inference, whereas Score communicates the same local model's completed inference as a highly task-specific scalar report.

Holding these payloads fixed, Fig.~\ref{fig:payload_reports}(b) shows the effect of the number of available reports. With one report, mean AUPRC is \(0.914\), \(0.925\), and \(0.933\) for TERRA-Score, TERRA-Feature, and Handcrafted-Stats, respectively. With three reports, these values increase to \(0.976\), \(0.978\), and \(0.966\), and with five reports they reach \(0.983\), \(0.985\), and \(0.969\). 
Across all evaluated report counts, TERRA-Feature slightly exceeds TERRA-Score in AUPRC, though the difference diminishes as more reports become available. The handcrafted representation performs best with a single report but saturates at a lower regional AUPRC as additional evidence is added. The largest gains for the two learned modes therefore arise from the first few additional reports, with comparatively modest improvement beyond \(K=5\).

\subsection{Robustness to Message-Recovery Errors}
\label{sec:results_ura}

\begin{figure*}[t]
    \centering
    \includegraphics[trim={0cm 0.3cm 0cm 0cm}, clip, width=0.9\textwidth]{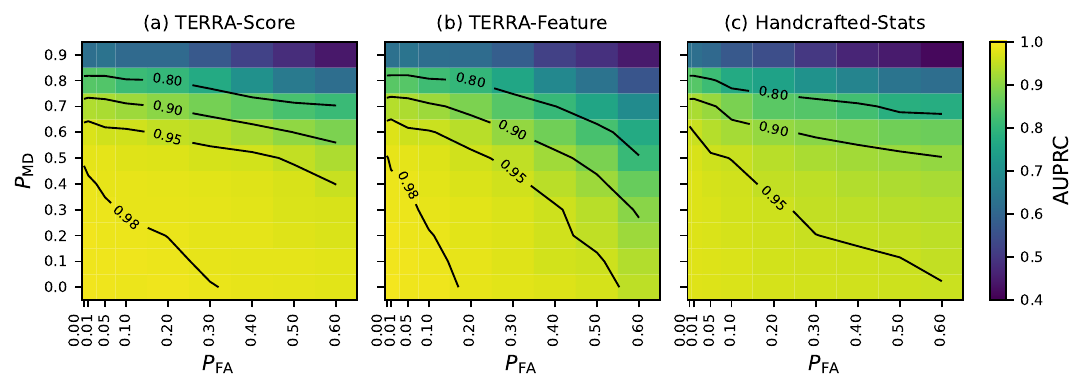}
    \caption{Regional fire-detection robustness to message-recovery errors for the selected reporting configurations with nine transmitted reports per region. Panels show mean AUPRC for (a) TERRA-Score, (b) TERRA-Feature, and (c) Handcrafted-Stats as a function of the global missed-detection probability \(P_{\mathrm{MD}}\) and false-alarm probability \(P_{\mathrm{FA}}\). All panels use the same color scale, and contours, where present, indicate AUPRC levels of 0.98, 0.95, 0.90, and 0.80. Metrics are averaged across 10 Monte Carlo recovery realizations within each outer fold, then averaged equally across the three folds.}
    \Description{Three heat maps showing regional AUPRC versus missed-detection and false-alarm probabilities for TERRA-Score, TERRA-Feature, and Handcrafted-Stats. Performance decreases with increasing recovery errors, while TERRA-Score is less sensitive to high false-message contamination over much of the evaluated region.}
    \label{fig:ura_robustness}
\end{figure*}

We next evaluate the selected reporting configurations over \(P_{\mathrm{MD}}\in[0,0.9]\) and \(P_{\mathrm{FA}}\in[0,0.6]\), starting from nine transmitted reports per region. Figure~\ref{fig:ura_robustness} reveals a different tradeoff from the perfect-recovery experiment. At \((P_{\mathrm{MD}},P_{\mathrm{FA}})=(0,0)\), TERRA-Feature has the highest mean AUPRC at \(0.9896\), followed by TERRA-Score at \(0.9868\) and Handcrafted-Stats at \(0.9713\). Under moderate joint impairment, however, TERRA-Score becomes more robust: at \((0.5,0.3)\), the corresponding AUPRCs are \(0.9605\), \(0.9414\), and \(0.9230\), and at \((0.7,0.4)\) they are \(0.8526\), \(0.8012\), and \(0.8173\). At the extreme corner \((0.9,0.6)\), where fewer than one genuine report remains per region on average, all methods degrade substantially, and TERRA-Feature is again slightly highest at \(0.4413\), compared with \(0.4383\) for TERRA-Score and \(0.4138\) for Handcrafted-Stats. Across the 90 sampled operating points, TERRA-Score has the highest aggregate mean AUPRC at 53 points, TERRA-Feature at 35, and Handcrafted-Stats at two; these counts characterize the sampled grid rather than implying uniform dominance by any method.

The two error mechanisms affect this ordering differently. When \(P_{\mathrm{FA}}=0\), missed messages mainly reduce the amount of genuine evidence, and TERRA-Feature preserves a small advantage over TERRA-Score through most of the evaluated range. For example, at \(P_{\mathrm{MD}}=0.5\), their mean AUPRCs are \(0.9812\) and \(0.9787\), respectively, compared with \(0.9635\) for Handcrafted-Stats, while at \(P_{\mathrm{MD}}=0.7\) the corresponding values are \(0.9290\), \(0.9255\), and \(0.9237\). Thus, the stronger robustness of TERRA-Score over much of Fig.~\ref{fig:ura_robustness} is not primarily a consequence of missing reports.

False decoded messages produce the clearer separation. With \(P_{\mathrm{MD}}=0\), TERRA-Feature remains slightly ahead at \(P_{\mathrm{FA}}=0.1\), with AUPRC \(0.9841\) compared with \(0.9831\) for TERRA-Score, but TERRA-Score overtakes it by \(P_{\mathrm{FA}}=0.2\), where the corresponding values are \(0.9838\) and \(0.9783\). At \(P_{\mathrm{FA}}=0.6\), TERRA-Score retains an AUPRC of \(0.9729\), compared with \(0.9398\) for TERRA-Feature and \(0.9512\) for Handcrafted-Stats. The low-false-alarm metric shows the same sensitivity: TERRA-Score and TERRA-Feature both begin at MDR \(1.81\%\) for FAR \(\leq1\%\), but at \(P_{\mathrm{MD}}=0\) and \(P_{\mathrm{FA}}=0.6\) the MDRs increase to \(5.34\%\) and \(16.56\%\), respectively, compared with \(13.35\%\) for Handcrafted-Stats.

A plausible explanation is that a false TERRA-Score report is constrained to a scalar in \([0,1]\) and then averaged with the genuine probabilities, whereas an independently generated false feature report can correspond to an atypical point relative to the learned feature distribution and is processed nonlinearly by the DeepSets receiver. This interpretation is consistent with the observed greater sensitivity of TERRA-Feature to false reports, but the present experiments do not establish it as a causal mechanism. More generally, the relative ranking is metric- and regime-dependent in the most severe conditions, so the results support a tradeoff rather than a universal ordering: TERRA-Feature provides the strongest regional discrimination when message recovery is reliable, while the more compact TERRA-Score is less sensitive to spurious decoded evidence over much of the evaluated impairment region.

The surfaces in Fig.~\ref{fig:ura_robustness} characterize downstream inference at a specified message-recovery quality rather than performance under equal physical-channel conditions. Because the reporting modes have different message lengths, a concrete access scheme may produce different \((P_{\mathrm{MD}},P_{\mathrm{FA}})\) operating points for TERRA-Score and TERRA-Feature at the same SNR, blocklength, and activity level. The present experiment deliberately holds these recovery probabilities fixed to isolate application-layer sensitivity; we discuss the implications for concrete low-power communication systems next.

\subsection{Communication Interpretation and Limitations}
\label{sec:discussion}

The selected operating points contain 6, 24, and 30 task bits per report for TERRA-Score, TERRA-Feature, and Handcrafted-Stats, respectively. These values exclude the region tag and nonce; the \(G=16\) address space requires at least four region-tag bits. Since the nonce length is implementation-dependent, we do not assign total message lengths or infer channel-use, energy, or latency reductions. In a concrete access scheme, message length, SNR, blocklength, activity, and coding jointly determine the achievable \((P_{\mathrm{MD}},P_{\mathrm{FA}})\), which can then be mapped through the robustness surfaces above to regional detection performance.

TERRA's task-oriented processing and regional aggregation can also operate over conventional addressed IoT links such as LoRaWAN \cite{Sasmita2018LoRaFire,Herring2022UndergroundLoRa}, which could transmit the same score or feature payloads and apply the same regional aggregation, but would not provide the identity-free reporting semantics of the main TERRA model. Those semantics require common-codebook transmission and list recovery (e.g., URA). TERRA therefore separates what task information is communicated and fused from how it is transported.

EN54 provides synchronized multisensor observations within one controlled indoor fire-test environment rather than multiple independently instrumented geographic regions or an outdoor wildfire deployment. The communication study accordingly embeds each EN54-derived target-region example in a synthetic \(G=16\) region-tag space and applies controlled list-level errors. Future work will connect these results to concrete low-power access schemes and larger outdoor deployments.

\section{Conclusion}
\label{sec:conclusion}

We introduced TERRA, a task-oriented distributed edge-AI architecture that couples compact local representations with identity-free region-tagged reporting and permutation-invariant regional evidence aggregation. We evaluated TERRA on the distributed indoor EN54 testbed for the episode-level fire-scenario target used in the regional evaluation. A four-dimensional TERRA-Feature report achieves mean AUPRC \(0.990\) using 24 task bits, while TERRA-Score achieves \(0.987\) using only 6 bits; both attain an MDR of \(1.81\%\) at FAR \(\leq1\%\). TERRA-Feature therefore achieves a small AUPRC advantage under reliable recovery, whereas TERRA-Score provides a substantially more compact task representation.

Under message-recovery errors, TERRA-Feature retains a small advantage when errors are dominated by missed reports, whereas TERRA-Score is less sensitive to false decoded messages over much of the evaluated region. Future work will map these application-level robustness results to concrete low-power access schemes and evaluate TERRA on larger outdoor deployments and richer regional inference tasks.

\begin{acks}
This research was supported by the National Edge AI Hub for Real Data (UK EPSRC, EP/Y028813/1), and the
Communications Hub for Empowering Distributed Cloud Computing Applications and Research (CHEDDAR) under grant numbers (UK EPSRC, EP/Y037421/1) and (UK EPSRC, EP/X040518/1).
\end{acks}

\bibliographystyle{ACM-Reference-Format}
\bibliography{references}

@inproceedings{Sasmita2018LoRaFire,
  author    = {Errynando Surya Sasmita and Mia Rosmiati and Moch Fahru Rizal},
  title     = {Integrating Forest Fire Detection with Wireless Sensor Network Based on Long Range Radio},
  booktitle = {Proceedings of the 2018 International Conference on Control, Electronics, Renewable Energy and Communications (ICCEREC)},
  pages     = {222--225},
  year      = {2018},
  publisher = {IEEE},
  doi       = {10.1109/ICCEREC.2018.8711991}
}

@inproceedings{Freira2024WildfireIoT,
  author    = {Daniel Filipe Gon{\c{c}}alves Freira and Carlos Coutinho},
  title     = {Wildfire Detection Based on {IoT} Technology},
  booktitle = {Proceedings of the 2024 8th International Symposium on Multidisciplinary Studies and Innovative Technologies (ISMSIT)},
  pages     = {1--6},
  year      = {2024},
  publisher = {IEEE},
  doi       = {10.1109/ISMSIT63511.2024.10757256}
}

@article{Herring2022UndergroundLoRa,
  author  = {Ben Herring and Tony Sharp and Tim Roberts and Jarred Fastier-Wooller and Greg Kelly and Oz Sahin and David Thiel and Dzung Dao and Peter L. Woodfield},
  title   = {Underground {LoRa} Sensor Node for Bushfire Monitoring},
  journal = {Fire Technology},
  volume  = {58},
  number  = {3},
  pages   = {1087--1095},
  year    = {2022},
  month   = {May},
  doi     = {10.1007/s10694-022-01224-3}
}

@article{Liu2020AirComp,
  author  = {Wanchun Liu and Xin Zang and Yonghui Li and Branka Vucetic},
  title   = {Over-the-Air Computation Systems: Optimization, Analysis and Scaling Laws},
  journal = {IEEE Transactions on Wireless Communications},
  volume  = {19},
  number  = {8},
  pages   = {5488--5502},
  year    = {2020},
  month   = {August},
  doi     = {10.1109/TWC.2020.2993703}
}

@article{Sahin2023AirCompSurvey,
  author  = {Alphan {\c{S}}ahin and Rui Yang},
  title   = {A Survey on Over-the-Air Computation},
  journal = {IEEE Communications Surveys \& Tutorials},
  volume  = {25},
  number  = {3},
  pages   = {1877--1908},
  year    = {2023},
  doi     = {10.1109/COMST.2023.3264649}
}

@article{Zhou2019EdgeIntelligence,
  author = {Zhi Zhou and Xu Chen and En Li and Liekang Zeng and Ke Luo and Junshan Zhang},
  title = {Edge Intelligence: Paving the Last Mile of Artificial Intelligence With Edge Computing},
  journal = {Proceedings of the IEEE},
  volume = {107},
  number = {8},
  pages = {1738--1762},
  year = {2019},
  month = {August},
  doi = {10.1109/JPROC.2019.2918951}
}

@inproceedings{Teerapittayanon2017DDNN,
  author = {Surat Teerapittayanon and Bradley McDanel and H. T. Kung},
  title = {Distributed Deep Neural Networks Over the Cloud, the Edge and End Devices},
  booktitle = {Proceedings of the 2017 IEEE 37th International Conference on Distributed Computing Systems (ICDCS)},
  pages = {328--339},
  year = {2017},
  publisher = {IEEE},
  doi = {10.1109/ICDCS.2017.226}
}

@article{Gunduz2023BeyondBits,
  author = {Deniz G{\"u}nd{\"u}z and Zhijin Qin and I{\~n}aki Estella Aguerri
            and Harpreet S. Dhillon and Zhaohui Yang and Aylin Yener
            and Kai Kit Wong and Chan-Byoung Chae},
  title = {Beyond Transmitting Bits: Context, Semantics, and
           Task-Oriented Communications},
  journal = {IEEE Journal on Selected Areas in Communications},
  volume = {41},
  number = {1},
  pages = {5--41},
  year = {2023},
  month = {January},
  doi = {10.1109/JSAC.2022.3223408}
}

@article{Shao2022TaskOriented,
  author = {Jiawei Shao and Yuyi Mao and Jun Zhang},
  title = {Learning Task-Oriented Communication for Edge Inference:
           An Information Bottleneck Approach},
  journal = {IEEE Journal on Selected Areas in Communications},
  volume = {40},
  number = {1},
  pages = {197--211},
  year = {2022},
  month = {January},
  doi = {10.1109/JSAC.2021.3126087}
}

@article{Chan2024WildfireIoTSurvey,
  author = {Chiu Chun Chan and Sheeraz A. Alvi and Xiangyun Zhou and Salman Durrani and Nicholas Wilson and Marta Yebra},
  title = {A Survey on {IoT} Ground Sensing Systems for Early Wildfire Detection: Technologies, Challenges, and Opportunities},
  journal = {IEEE Access},
  volume = {12},
  pages = {172785--172819},
  year = {2024},
  doi = {10.1109/ACCESS.2024.3501336}
}

@article{Vorwerk2023Distance,
  author = {Pascal Vorwerk and J{\"o}rg Kelleter and Steffen M{\"u}ller and Ulrich Krause},
  title = {Distance-Based Analysis of Early Fire Indicators on a New Indoor Laboratory Dataset with Distributed Multi-Sensor Nodes},
  journal = {Fire},
  volume = {6},
  number = {8},
  pages = {323},
  year = {2023},
  doi = {10.3390/fire6080323}
}

@misc{Vorwerk2025Dataset,
  author       = {Pascal Vorwerk},
  title        = {Indoor Fire Dataset with Distributed Multi-Sensor Nodes (EN54 Test Room)},
  year         = {2025},
  howpublished = {Mendeley Data, Version 2},
  doi          = {10.17632/npk2zcm85h.2}
}

@ARTICLE{Ozates2026URASurvey,
  author={Ozates, Mert and Javad Ahmadi, Mohammad and Kazemi, Mohammad and Gündüz, Deniz and Duman, Tolga M.},
  journal={IEEE Communications Surveys \& Tutorials}, 
  title={Unsourced Random Access: A Comprehensive Survey}, 
  year={2026},
  volume={28},
  number={},
  pages={955-984},
  doi={10.1109/COMST.2025.3637685}}

@inproceedings{Zaheer2017DeepSets,
  author    = {Manzil Zaheer and Satwik Kottur and Siamak Ravanbakhsh and Barnabas P{\'o}czos and Ruslan Salakhutdinov and Alexander J. Smola},
  title     = {Deep Sets},
  booktitle = {Advances in Neural Information Processing Systems},
  volume    = {30},
  pages     = {3391--3401},
  year      = {2017}
}

\end{document}